\documentclass[final]{l4dc2026}

\title[LINDI]{Learned Incremental Nonlinear Dynamic Inversion for Quadrotors with and without Slung Payloads}
\usepackage{times}
\usepackage{graphicx}
\usepackage{amssymb}
\usepackage{amsmath}
\usepackage{silence}
\usepackage{caption}
\usepackage{siunitx}

\newcommand{\vq}{\mathbf{q}}
\newcommand{\vqhat}{\hat{\mathbf{q}}}
\newcommand{\vp}{\mathbf{p}}

\newcommand{\ve}{\mathbf{e}}
\newcommand{\vv}{\mathbf{v}}

\newcommand{\vx}{\mathbf{x}}

\newcommand{\vu}{\mathbf{u}}

\newcommand{\vf}{\mathbf{f}}
\newcommand{\vomega}{\boldsymbol{\omega}}

\newcommand{\vtau}{\boldsymbol{\tau}}

\newcommand{\mR}{\mathbf{R}}

\newcommand{\mK}{\mathbf{K}}

\newcommand{\mJ}{\mathbf{J}}

\DeclareMathOperator*{\argmin}{argmin}

\newcommand{\highlight}[1]{%
  \colorbox{green!50}{$\displaystyle#1$}}

\author{%
  \Name{Eckart Cobo-Briesewitz}$^1$ \Email{cobo-briesewitz@campus.tu-berlin.de}\\
  \Name{Khaled Wahba}$^1$ \Email{k.wahba@tu-berlin.de}\\
  \Name{Wolfgang Hönig}$^{1,2}$ \Email{hoenig@tu-berlin.de}\\
  \addr $^1$Technical University of Berlin, Berlin, Germany\\
  \addr $^2$Robotics Institute Germany (RIG)
}

\begin{document}
 \sisetup{mode=match}

\maketitle

\begin{abstract}
The increasing complexity of multirotor applications demands flight controllers that can accurately account for all forces acting on the vehicle. Conventional controllers model most aerodynamic and dynamic effects but often neglect higher-order forces, as their accurate estimation is computationally expensive. Incremental Nonlinear Dynamic Inversion (INDI) offers an alternative by estimating residual forces from differences in sensor measurements; however, its reliance on specialized and often noisy sensors limits its applicability. Recent work has demonstrated that residual forces can be predicted using learning-based methods. In this paper, we show that a neural network can generate smooth approximations of INDI outputs without requiring specialized rotor RPM sensor inputs. We further propose a hybrid approach that integrates learning-based predictions with INDI and demonstrate both methods for multirotors and multirotors carrying slung payloads. Experimental results on trajectory tracking errors demonstrate that the specialized sensor measurements required by INDI can be eliminated by replacing the residual computation with a neural network.
\end{abstract}


\section{Introduction}
The rise in complexity of tasks to be executed by multirotors has necessitated the development of more accurate flight controllers that are able to model all forces acting on these robots. Traditional flight controllers \citep{lee2010} model a large part of these forces but are unable to model all of them. The remaining forces, called residual forces, can arise from diverse sources such as blade flapping, drag, or ground forces \citep{shi2020, 2022-shi-NeuralSwarm2PlanningControl, bauersfeld2021}. Trying to compute residual forces directly can be computationally complex and intractable for real time control, especially on quadrotors with restricted hardware.

Recent work has shown that learning algorithms can lead to significant improvements in flight performance \citep{bauersfeld2021}, even modeling the interaction forces between multiple quadrotors \citep{shi2020}. By only learning the residual forces rather than the full dynamics of the quadrotor, the amount of training data needed is greatly reduced. In addition, learning models can help improve flight performance when sensor readings are noisy or lacking.


Incremental Nonlinear Dynamic Inversion (INDI) \citep{2016-smeur-AdaptiveIncrementalNonlinear} derives the residual forces from the mismatch between the nominal dynamic model and the real-time sensor feedback, using this residual to incrementally refine the control input. In the context of Unmanned Aerial Vehicles (UAVs), INDI typically relies on rotor RPM measurements to approximate the nominal model. Although INDI can significantly enhance flight performance, its effectiveness depends on the availability of specific sensor data and may still be affected by measurement noise.

One promising application of multirotors is payload transportation, particularly in time-sensitive fields such as medicine and rescue operations. When robots carry a payload attached by a cable \citep{tang2014aggressive, Wahba2023CableForceAllocation}, the dynamics of the multirotor become more complex due to the inclusion of the payload, which introduces additional forces on the system. These added dynamics can give rise to new types of unmodeled residual forces, thereby increasing the need for accurate estimators and making residual force prediction even more crucial.


\begin{figure}[htbp]
  \centering
  \includegraphics[width=0.8\linewidth]{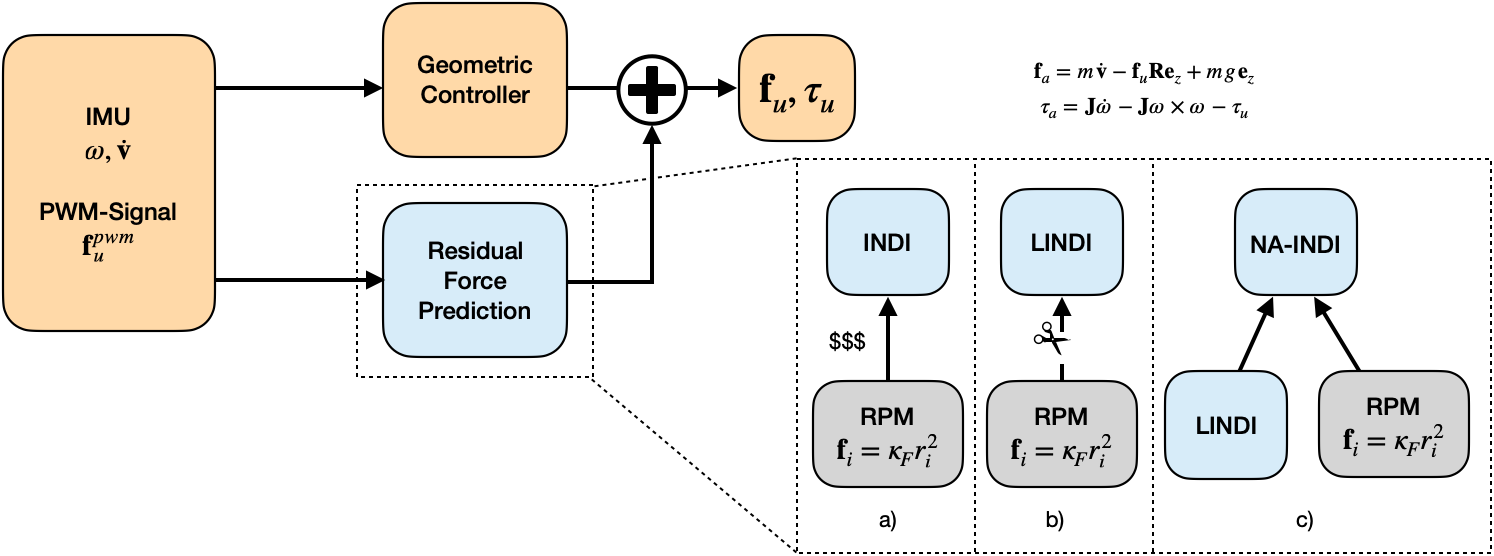}
  \caption{Overview of the three approaches for computing residual forces compared in our paper: a) Standard INDI relies on additional sensor measurements, which can be expensive and prone to noise. b) Learned INDI (LINDI) learns the outputs of INDI, eliminating the need for specialized sensors and yielding smoother predictions. c) Neural Augmented INDI (NA-INDI) fuses LINDI predictions with the original sensor data, resulting in higher-quality residual force estimates. LINDI and NA-INDI are formally defined in Section 4.}
  \label{fig:visabs}
\end{figure}

We propose Learned Incremental Nonlinear Dynamic Inversion (LINDI) a neural-network-based approach that achieves comparable flight performance to the INDI controller without requiring rotor RPM measurements. By replacing the incremental model with a learned component integrated into two traditional flight controllers \citep{lee2010, yu2020geometric}, our approach removes the need for specialized RPM sensors. Quantitative results show that it matches INDI’s performance, and we further analyze a hybrid setup, Neural-Augmented Incremental Nonlinear Dynamic Inversion (NA-INDI), combining both INDI and neural network predictions. Figure~\ref{fig:visabs} summarizes the three evaluated approaches.

\begin{figure}[htbp]
    \centering
    \includegraphics[width=0.3\textwidth]{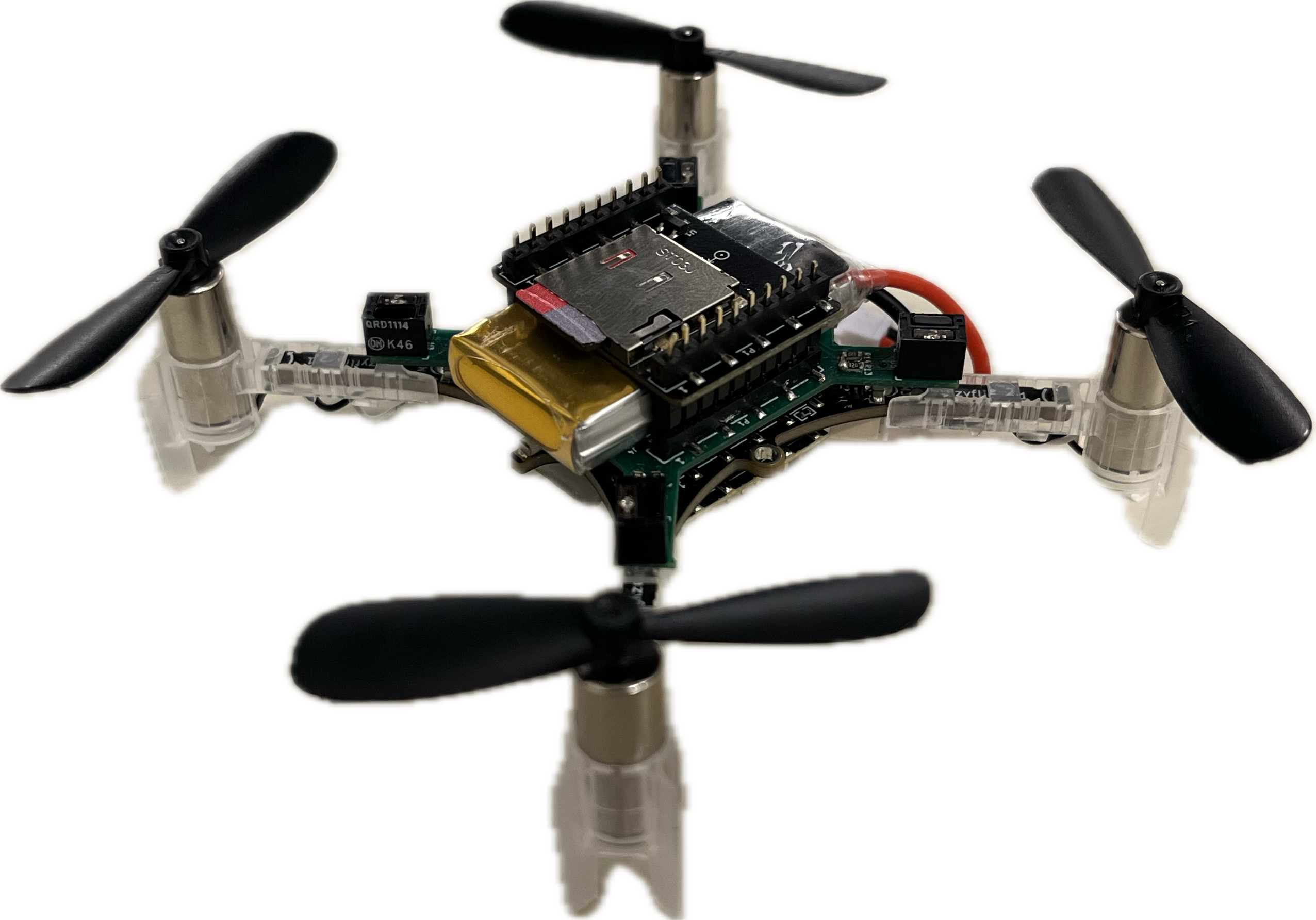}
    \centering
    \includegraphics[width=0.08\textwidth]{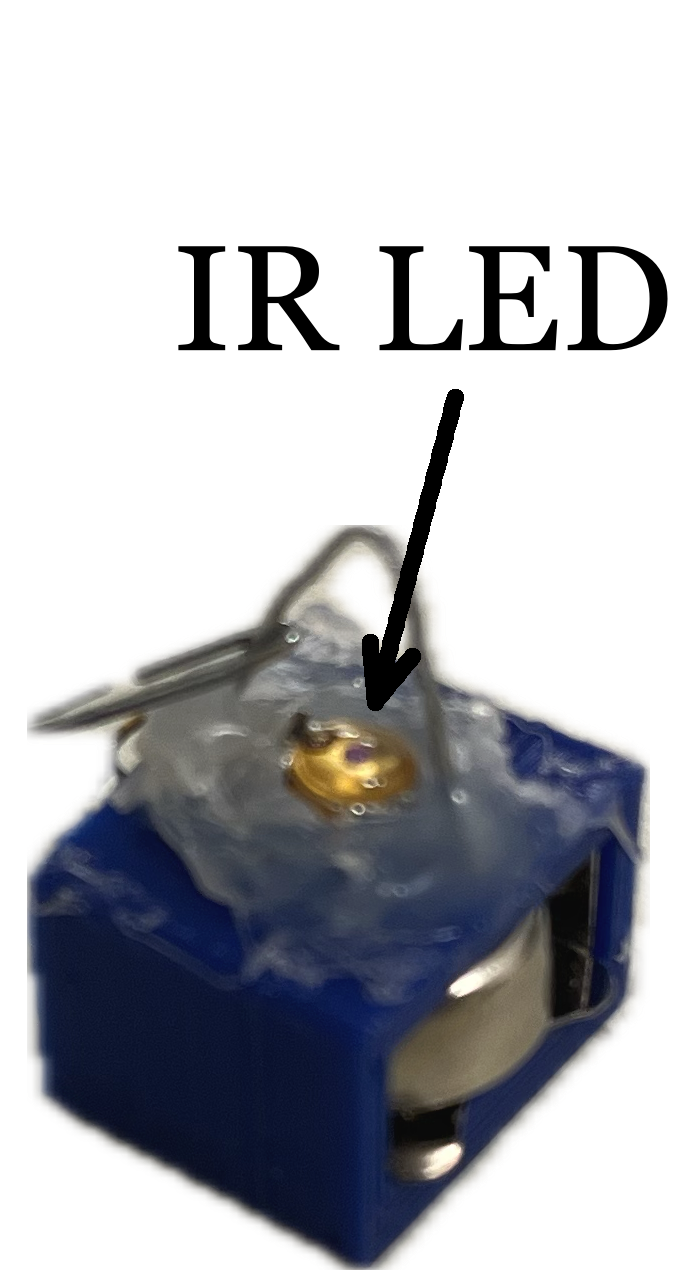}
    \centering
    \caption{Hardware used for experiments. We rely on a Bitcraze Crazyflie 2.1 with standard motors, Micro-SD-card extension, and custom RPM measurement board that uses IR-LEDs. These LEDs can also be tracked by a motion capture system. The payload is equipped with another active IR LED for tracking.}
    \label{fig:hardware}
\end{figure}

\section{Related Work}
Several prior studies have highlighted the importance of INDI for achieving high tracking performance, particularly in high-speed flight. Specifically, incorporating the INDI component is essential for both MPC- and geometric-based controllers, with both approaches achieving comparable performance when provided with a valid reference trajectory~\citep{Sun2022QuadrotorMPC}. Performance can be further enhanced by modeling drag~\citep{Sun2022QuadrotorMPC, 2018-faessler-DifferentialFlatnessQuadrotor} or motor delay~\citep{2021-tal-AccurateTrackingAggressive}, though these factors have a comparatively smaller effect. More recently, INDI has also been shown to play a critical role in agile flight of multirotors carrying payloads~\citep{2025-sun-AgileCooperativeAerial}.

An alternative to online estimation, as in INDI, is to learn a function that directly predicts the nonlinear residual dynamics. \citet{bauersfeld2021} propose augmenting the controller with a neural network that predicts residual forces acting on a quadrotor. Their method adopts a non-Markovian assumption, using previous states as inputs rather than only the current state. The network outputs a six-dimensional vector representing the residual forces and torques along the $x$-, $y$-, and $z$-axes. Experimental results demonstrate that such learned residual models can significantly enhance flight performance. Similarly, \citet{2022-shi-NeuralSwarm2PlanningControl} investigate residual forces arising from inter-quadrotor interactions, particularly the downwash effect, where an upper quadrotor induces downward forces on a lower one. Their approach employs DeepSets to aggregate outputs from individual neural networks for each quadrotor, enabling the prediction of residual forces specifically along the $z$-axis. This method reduces the worst-case $z$-position error under downwash conditions by a factor of two to four.

Recent studies have explored integrating the INDI framework with learning-based models. \citet{9837175} employ Gaussian Processes to correct sensor measurement inaccuracies in real time; their method updates the Gaussian Process online. In contrast, \citet{Zhang2023MetaINDI} use meta-learning to continuously adapt residual predictions during flight, updating a neural network to estimate the INDI control effectiveness matrix. Our approach differs from previous work in that the model is trained offline and directly predicts INDI outputs, avoiding the instability that may occur when online models are not fully converged and removing the need for special sensor measurements. Additionally, unlike prior work, we validate our method through extensive real world evaluations on two different quadrotor modalities.

There is also prior work combining learning-based methods with slung-type payload transport in multirotors. \citet{Jin2024NeuralPredictor}, for instance, employ a neural network to predict forces on a multirotor caused by an unknown payload. In contrast, our work assumes a known payload and focuses exclusively on residual dynamics.

The main contribution of our approach relative to previous works is that it eliminates the need for rotor RPM measurements while empirically demonstrating flight performance comparable to INDI, with smoother and less noisy residual force predictions.

\section{Problem Description}
We first introduce the dynamics that we consider in this work, followed by the control problems that we try to solve.

\subsection{Multirotor Dynamics}
Consider a multirotor modeled as a rigid floating base with state $\vx = [\vp, \mR, \vv, \vomega]^\top$. 
Here, $\vp, \vv \in \mathbb{R}^3$ represent the position and velocity in the world frame, $\mR \in SO(3)$ represents the rotation matrix from body to world, and $\vomega \in \mathbb{R}^3$ is the angular velocity expressed in the body frame. 
The action $\mathbf{u} \in \mathbb{R}^4$ is defined as the angular velocities of the rotors, $\vu = [r_{1}, r_{2}, r_{3}, r_{4}]^\top$. 
The dynamics are derived from Newton-Euler equations for rigid bodies as
\begin{align}
    &\dot{\vp} = \vv, && m\dot{\mathbf{v}} = f_u \mathbf{R} \ve_z - m g \ve_z + \vf_a,\\
    &\dot{\mR} = \mR\hat{\vomega}, &&
    \mJ\dot{\vomega} = \mJ\vomega\times \vomega + \vtau_u + \vtau_a \label{eq:dynamics},
\end{align}
where $m$ is the mass, $\mathbf{J}$ is the inertia matrix, $g$ is the gravitational acceleration constant, $\ve_z = [0, 0, 1]^\top$, $f_u$ is the collective thrust created by the rotors, $\vtau_u$ is the torque vector created by the rotors, $\vf_a, \vtau_a$ are unknown external forces and torques, respectively, and $\hat{\vomega}$ denotes the skew-symmetric matrix of $\vomega$ such that $\hat{\vomega}\mathbf{v} = \vomega \times \mathbf{v}$ for any $\mathbf{v}$.

The relationship between the motor angular velocity and the generated total force and torque on the rigid body is
\begin{subequations}
\label{eq:motors}
\begin{align}
f_i &= \kappa_F r_i^2,\\
[f_u, \vtau_u]^\top &= \mathbf{B}_0 [f_1, f_2, f_3, f_4]^\top,
\end{align}
\end{subequations}
where $\kappa_F$ is a propeller constant and $\mathbf{B}_0$ is a fixed and known actuation matrix.

\subsection{Multirotor With Cable-Suspended Payload}

Consider a slung type payload attached to the center of gravity of the multirotor. 
The multirotors state is extended to include the payload position $\vp_p$ and velocity $\vv_p$.
The translational dynamics are extended with additional terms:
\begin{align}
\label{eq:payloadyn}
m\dot{\mathbf{v}} = f_u \mathbf{R} \ve_z - m g \ve_z \highlight{+ T \vq} + \vf_a, \\
\dot{\vp}_p = \vv_p, \quad m_p \dot{\vv}_p = -T\mathbf{q} - m_p g\ve_z 
\end{align}
where $\vq = \frac{\vp_p - \vp}{\| \vp_p - \vp \| }$ indicates the direction of the cable, $m_p$ is the mass of the payload, and $T$ is the cable tension.
From \eqref{eq:payloadyn}, the tension can be computed as
\begin{align}
\label{eq:tension}
T =  -m_p \mathbf{q}^\top (\dot{\vv}_p +  g\ve_z).
\end{align}

\subsection{Control Problem}

Given the nominal models and feasible reference trajectories $\vx_r(t)$, we are aiming to find controllers that can minimize the mean tracking error:
\begin{align}
\argmin_\pi \frac{1}{D} \int_{t=0}^D d(\vx_\pi(t), \vx_r(t)) dt, \label{eq:trackinError}
\end{align}
where $\pi(\vx, \vx_r) \mapsto \vu$ is the control law that influences the state $\vx_\pi(t)$, $D$ the duration of the reference trajectory $\vx_r(t)$, and $d$ is a distance metric.
For the multirotor case, we consider the positional tracking error, i.e., $d(\vx_\pi, \vx_r)=\|\vp_\pi - \vp_r\|_2$ and for the payload transport case the positional tracking error of the payload. 

In this work, we augment classical controllers that ignore $\vf_a$ and $\vtau_a$ with measured and/or predicted components to improve the tracking performance \eqref{eq:trackinError}.

\section{Approach}
This section introduces our two proposed methods. Learned Incremental Nonlinear Dynamic Inversion (LINDI) is a learning-based approach that eliminates the need for the additional sensor measurements required by INDI, while maintaining --- or even improving --- performance by learning from smoothed data. Neural-Augmented Incremental Nonlinear Dynamic Inversion (NA-INDI) combines the outputs of a neural network with INDI sensor measurements to achieve performance superior to both INDI and LINDI.

\subsection{Geometric Control Laws} \label{sub:controllers}

\subsubsection{Multirotor}
An exponentially stable geometric controller for a multirotor computes the desired force and torques as follows~\citep{lee2010}:
\begin{subequations}
    \label{eq:controller}
    \begin{align}
        \begin{split}
        f_u &= (-\mK_p \ve_p - \mK_v \ve_v + m g \ve_z + m \ddot{\vp}_d 
        \highlight{-\vf_a}) \cdot \mR \ve_z,
        \end{split}
        \\
        \label{eq:controller:b} 
        \begin{split}
        \vtau_u &= - \mK_R \ve_R - \mK_\omega \ve_\omega - \mJ\vomega\times \vomega \\ 
        &\qquad - \mJ(\hat\vomega \mR^\top \mR_d \vomega_d - \mR^\top \mR_d \dot{\vomega}_d) \highlight{-\vtau_a},
        \end{split}
    \end{align}
\end{subequations}
where $\mR_d$, $\vomega_d$ and $\dot{\vomega}_d$ are desired references that can be computed using differential flatness;  $\ve_p$, $\ve_v$, $\ve_R$, $\ve_\omega \in \mathbb{R}^3$ are errors with respect to this reference (mathematically defined in~\cite{lee2010}), and $\mK_p$, $\mK_v$, $\mK_R$, $\mK_\omega \in \mathbb{R}^{3\times 3}$ are diagonal positive gain matrices.

The highlighted parts $\vf_a$, $\vtau_a$ are new terms that need to be added to compensate for residual forces and torques.

\subsubsection{Multirotor with cable-suspended payload}
The previous control law can be extended to track a cable-suspended load~\citep{yu2020geometric}.
The controller is a cascaded design, where the first level tracks the position and velocity of the payload, the second level tracks the desired cable direction and its derivative, and the third level tracks the UAVs rotation and angular velocity.
Eventually, $f_u$ computes the control force generated by the multirotor to track the desired cable and payload motion, while attenuating the external residuals $\mathbf{f}_a$
\begin{align}
\label{eq:conroller2}
f_u &= (\vu_0 \highlight{-\vf_a}) \cdot \mR \ve_z,
\end{align}
where $\vu_0$ is the nominal control force to track the cable as defined in~\citet{yu2020geometric}.
Finally the rotational dynamics are identical to \eqref{eq:controller:b}, since the cable is assumed to be attached at the center of gravity and the mismatches between the nominal and real model are compensated using $\boldsymbol{\tau}_a$.

\subsection{Incremental Nonlinear Dynamic Inversion (INDI)}

The key idea of the Incremental Nonlinear Dynamic Inversion (INDI) is to estimate $\vf_a$ and $\vtau_a$ in real-time using IMU and RPM sensor measurements. The INDI controller is implemented on the multirotor dynamics in \citet{2016-smeur-AdaptiveIncrementalNonlinear,2021-tal-AccurateTrackingAggressive}. 
The dynamics of the multirotor-payload case differ from the standard case by the force applied by the cable. Therefore, re-arranging \eqref{eq:dynamics} and \eqref{eq:payloadyn} yields
\begin{subequations}
\label{eq:faTaua}
\begin{align}
\vf_a &= m\mathbf{\dot{v}} - f_u \mathbf{R} \ve_z + m g \ve_z - T \vq,\\
\vtau_a &= \mJ\dot{\vomega} - \mJ\vomega\times \vomega - \vtau_u.
\end{align}
\end{subequations}

In INDI, $f_u$ and $\vtau_u$ are computed from RPM measurements by applying \eqref{eq:motors}, $\dot \vv$ is measured by the accelerometer in body frame and rotated to world frame, and $\vomega$ is measured by the gyroscope.
Other values like $\dot \vomega$ and $\dot \vv_p$ (needed through \eqref{eq:tension})  are estimated numerically, while $\mR$ and $\vq$ are computed by the state estimation.
For practical purposes, it is important to filter the measured values, e.g., by using a butterworth filter~\citep{2021-tal-AccurateTrackingAggressive}.
The online estimated values can be used as additional feedforward terms in the controllers, as highlighted in \eqref{eq:controller} and \eqref{eq:conroller2}.

Within the INDI framework, multirotor–payload dynamics can be modeled either by explicitly including payload and cable dynamics~\citep{sreenath2013geometric1uav, yu2020geometric, li2023rotortm}, or by capturing the coupling via cable tension~\citep{2025-sun-AgileCooperativeAerial, sreenath2013dynamics}. We adopt the latter formulation in \eqref{eq:payloadyn} due to its practical suitability. The full formulation with explicit cable dynamics and the motivation for this choice are provided in the appendix.

\subsection{Learned Incremental Nonlinear Dynamic Inversion (LINDI)}
\label{sec:lndi}

Using a dataset containing example trajectories with IMU data, RPM data, and state estimates, one can also compute $\vf_a$ and $\vtau_a$ using \eqref{eq:faTaua}. Here, noisy data such $\dot \vv$ and $\dot \vomega$ can be pre-processed with zero-delay filters such as spline fitting.
The resulting values of $\vf_a$ and $\vtau_a$ are the labels of a supervised learning problem.

In the single quadrotor case, we use a multi-layer perceptron (MLP) with 19 inputs, 6 outputs, 2 hidden layers with 24 dimensions each, and Leaky-ReLU activation.
The input includes $\vv$, $\dot \vv$, $\vomega$, the first two columns of the rotation matrix $\mR$, and the motor PWM signal (which is different from $r_i$ as it cannot observe motor delays).
The output is the residual force and torque $[\vf_a, \vtau_a]^\top \in \mathbb R^6$.
The inputs and outputs of the network are scaled using min-max normalization to the range \([-1, 1]\) to mitigate disparities in value magnitudes and enhance training stability.

For the payload network the input is expanded to include the payload velocity, acceleration and cable direction, resulting in 28 inputs. These quantities are included as they provide relevant information for estimating the residual forces acting on the payload. The architecture remains the same except for the hidden layers which are reduced to a dimension of 16. This is due to the residual forces in the payload scenario being more pronounced and making the training process easier allowing for a smaller network size.

\begin{figure}[t]
    \centering
    \begin{minipage}[t]{0.3\textwidth}
        \includegraphics[width=\textwidth]{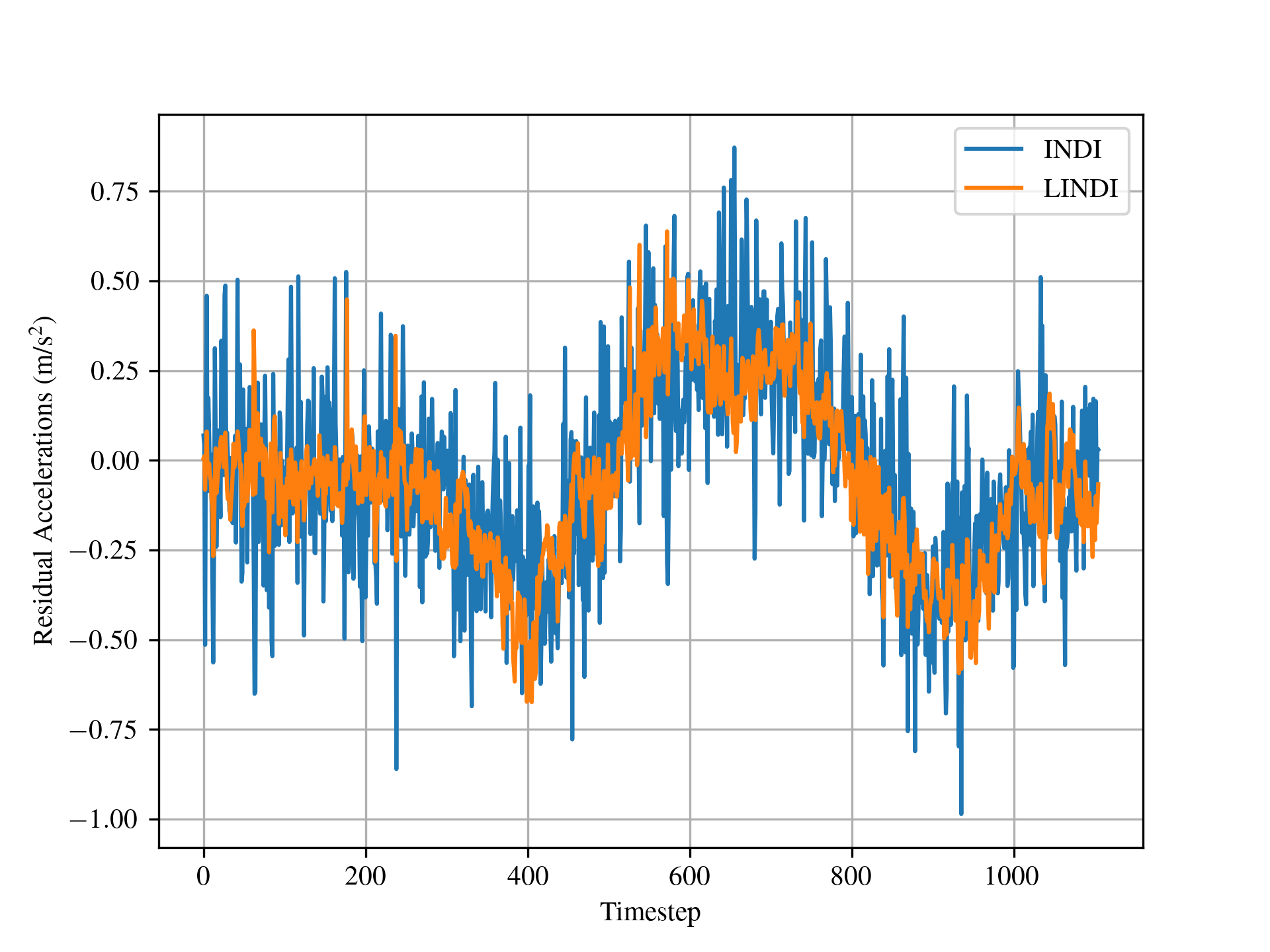}
    \end{minipage}
    \hspace{1cm}
    \begin{minipage}[t]{0.3\textwidth}
        \includegraphics[width=\textwidth]{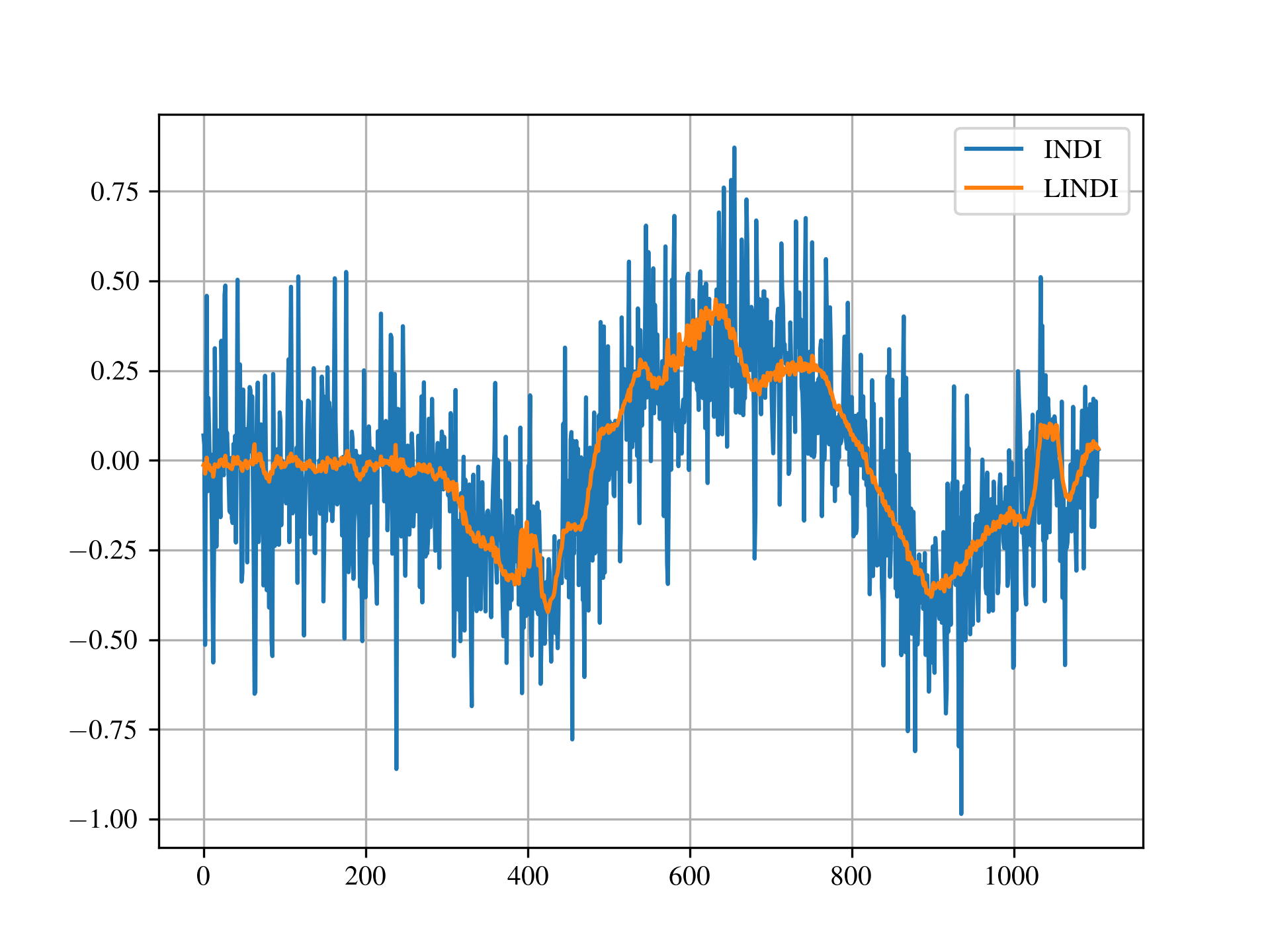}
    \end{minipage}
    \hfill
    \caption{The figure shows the outputs of two different MLPs for the residual accelerations along the y-axis, compared with the original INDI predictions during a Figure8 flight path for a quadrotor carrying a payload. The left plot presents the predictions of an MLP trained on the raw INDI outputs, while the right plot shows the results from an MLP trained on spline-fitted INDI outputs. Although the standard MLP exhibits some inherent de-noising capability, it produces noticeably smoother predictions when trained on less noisy data, especially when using a smaller dataset for training.}
    \label{fig:splines}
\end{figure}

For preprocessing, we fit splines composed of cubic polynomial segments that minimize the $L_2$ error with respect to the data points, rather than interpolating them exactly. These splines are applied to the collected INDI outputs, and the resulting smooth signals are used as training labels. The impact of training on smoothed data is illustrated in Figure \ref{fig:splines}. Because this smoothing is performed offline, it is more effective than the Butterworth filter used online in INDI.

\subsection{Neural-Augmented Incremental Nonlinear Dynamic Inversion (NA-INDI)}

We split the unmodeled dynamics in two parts $\vf_a=\vf_{a,NN} + \vf_{a,INDI}$ and similar for $\vtau_a$, where $\vf_{a,NN}$ and $\vtau_{a,NN}$ are learned functions that were trained using the steps described in \ref{sec:lndi}. Then the INDI control law only needs to reason about the remaining mismatch in the dynamics.


\begin{figure*}[t]
    \begin{minipage}[t]{0.22\textwidth}
        \centering
        \includegraphics[width=\textwidth]{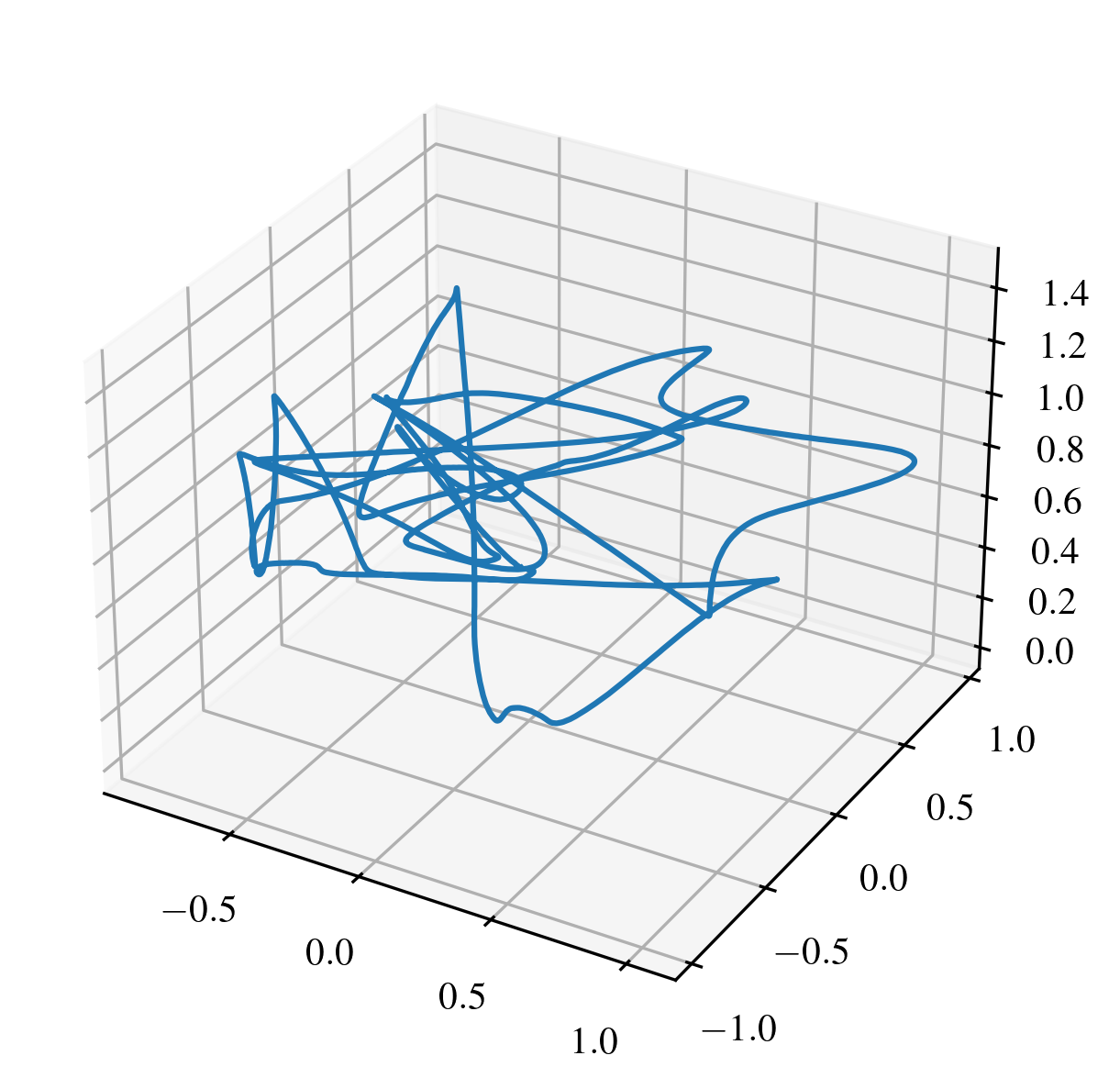}
        \caption*{(a) Random Points}
    \end{minipage}
    \centering
    \begin{minipage}[t]{0.22\textwidth}
        \centering
        \includegraphics[width=\textwidth]{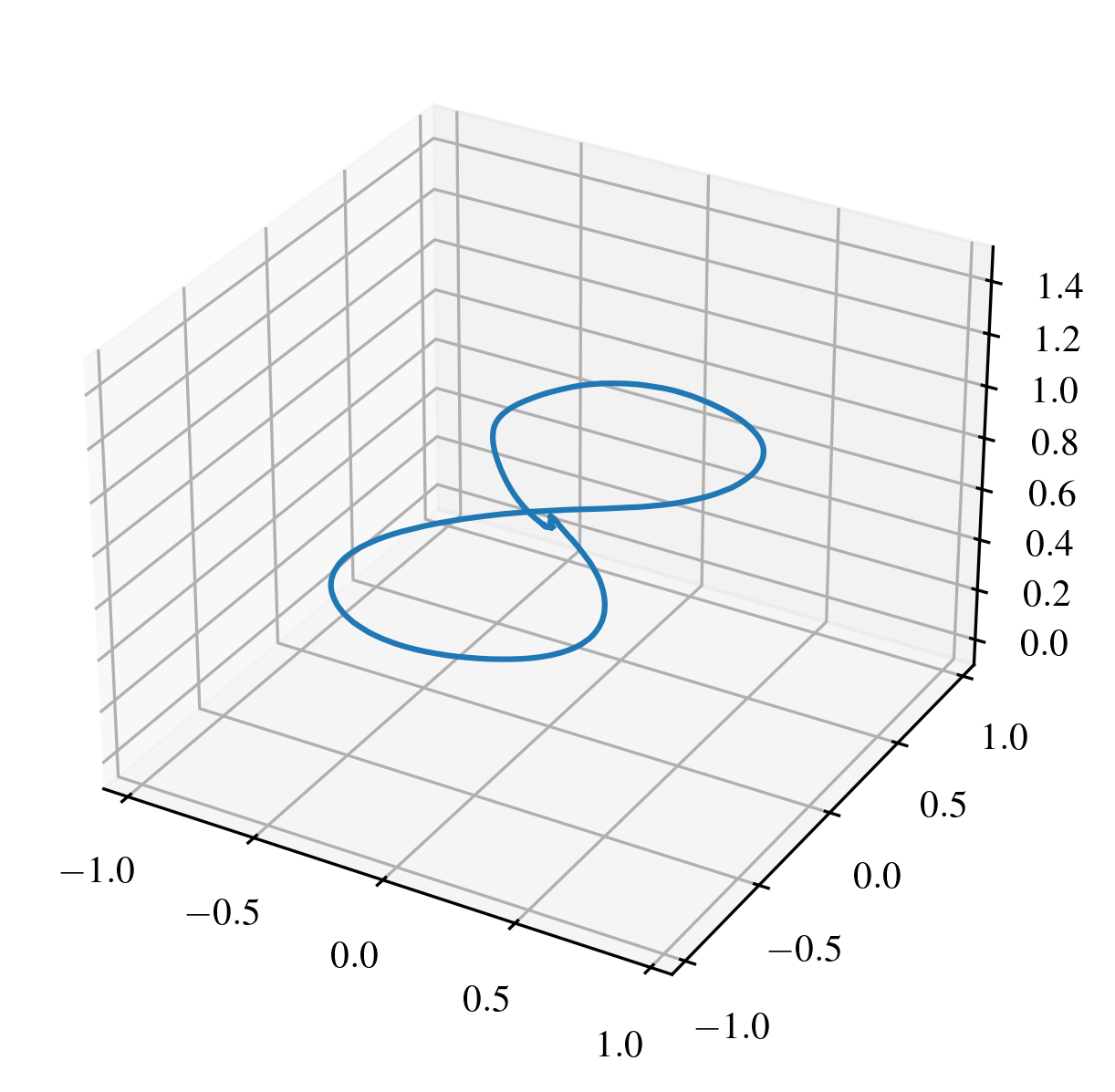}
        \caption*{(b) Figure8}
    \end{minipage}
    \centering
    \begin{minipage}[t]{0.22\textwidth}
        \centering
        \includegraphics[width=\textwidth]{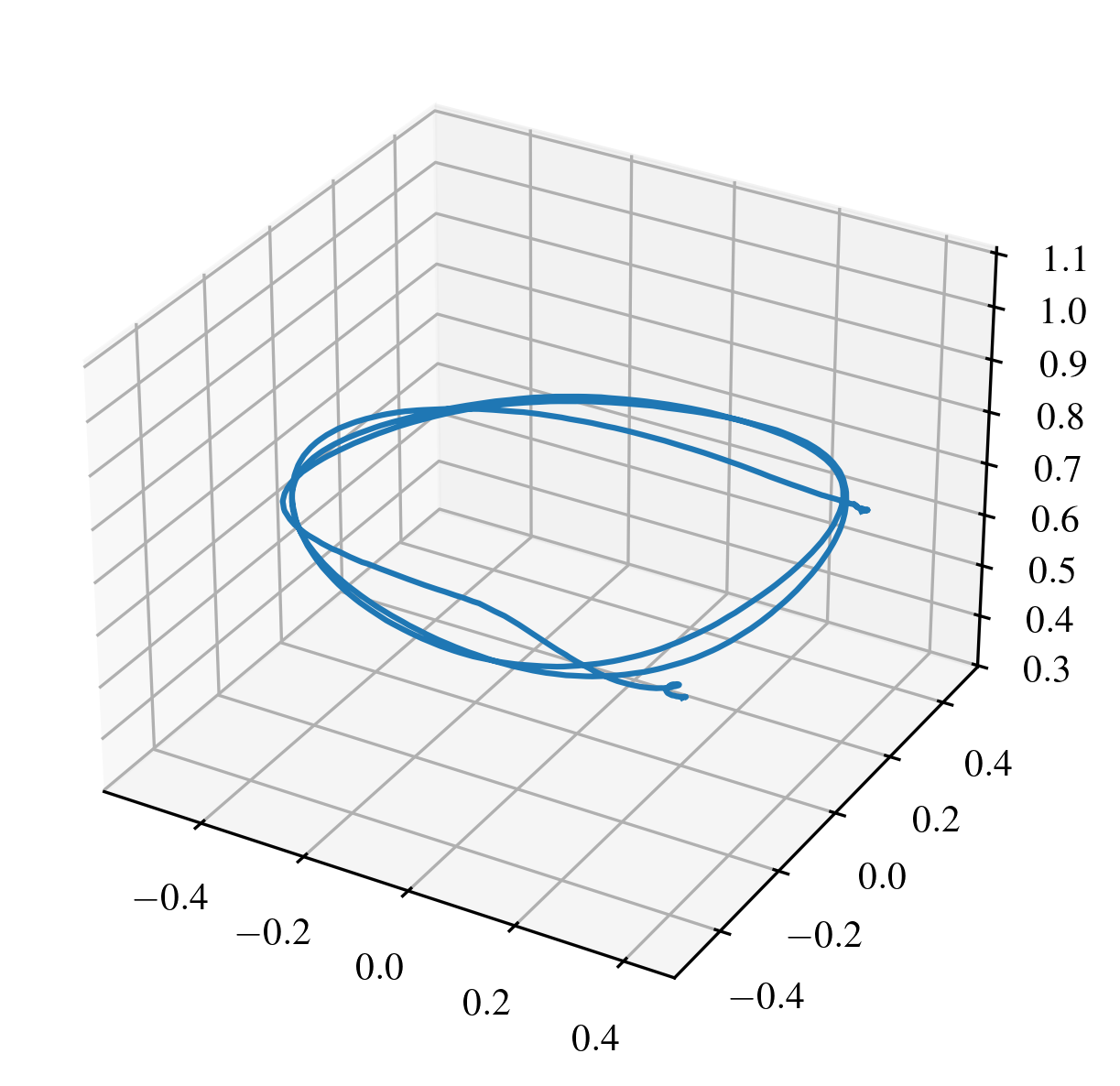}
        \caption*{(c) Circle}
    \end{minipage}
    \caption{The first image (a) shows a sample trajectory used to generate training data for the neural network. Random points are sampled within a predefined bounding box to define a spline trajectory, which the quadrotor follows at a speed of up to 2 m/s. Images (b) and (c) show the trajectories used for testing the performance of the different controllers.}
    \label{fig:trajectories}
\end{figure*}

\section{Experiments}
In order to quantify the performance of the different approaches, we perform real world flights on two unseen flight trajectories which are not present in the dataset used to train the neural network.

\subsection{Physical Setup}
We use the Bitcraze Crazyflie 2.1 quadrotors for our experiments, a small quadrotor with an arm length of \SI{4.6}{cm} and weight of \SI{34.7}{g}.
It is operated by an STM32 microprocessor (168 MHz, 1 MB of flash, and 192 KB RAM).
We equip the standard robot with two additions: a commercially available PCB to log data on a microSD card\footnote{https://www.bitcraze.io/products/micro-sd-card-deck/}, and a custom PCB to measure RPM as can be seen in Figure \ref{fig:hardware}.
In order to accurately predict the position of the quadrotor, we use an OptiTrack motion capture system running at 100 Hz in a \qtyproduct{7.5 x 4 x 2.75}{m} flight space.
The custom PCB to measure RPMs uses infrared emitter/receivers and can be used as an active marker by the Optitrack motion capture system to estimate the robot's pose.
Rotors have reflective markers underneath the blade, which allow the IR receiver to count the rotor rotations.
For communicating with the quadrotor we use Crazyswarm2 which is based on Crazyswarm \citep{Preiss2017Crazyswarm} but uses ROS 2 \citep{Macenski2022ROS2}. 

For the payload scenario we use a \SI{5}{g} weight attached at the center of the quadrotor by a string with \SI{0.5}{m} length. The string is connected to the quadrotor by a magnet. The payload is also equipped with an infrared LED for estimating its position as can be seen in Figure \ref{fig:hardware}. 
For simplicity, we keep the payload size and mass fixed, and refer the reader to prior work on adapting to varying payload parameters \citet{Jin2024NeuralPredictor}.

\subsection{Data Collection and Training}
We train our neural networks on a set of trajectories generated by sampling random points within a predefined bounding box of dimensions $1.6 \times 1.6 \times 0.4,\mathrm{m}^3$. These points are used to define spline trajectories, which the quadrotor follows at a randomly sampled velocity of up to 2 m/s in the no-payload case and up to 1.75 m/s in the payload case.

We collected a total of 64,996 timestamps, or around 12 minutes of flight data, for the no-payload scenario each one being a pair of state and residual force and torque, and  58,538 timestamps for the payload scenario.

We use the ADAM optimizer \citep{kingma2014adam} with an $L_1$-loss and with an initial learning rate of $3 \cdot 10^{-1}$ while reducing the learning rate by a factor of 0.92 every 10 epochs to enhance training stability and convergence, training for a total of 128 epochs with a batch size of 512. The networks for the single UAV and the UAV with a payload are trained using the same procedure but with datasets corresponding to their respective scenarios. 
We found similar performance across hyperparameter settings in offline testing, so we omit a detailed ablation study. Our goal is to show that learning-based models can replace specialized sensor measurements, not to optimize a specific architecture.


We run all neural networks locally on the quadrotor hardware. For the MLP described in section \ref{sec:lndi}, the inference time is 203 microseconds on the Crazyflie hardware.

\subsection{Testing Scenarios}
We test the performance of the controllers on two different predefined trajectories which are not present in the training set. The trajectories can be seen in Figure \ref{fig:trajectories}.
We compute the mean tracking error \eqref{eq:trackinError} per trial and repeat over multiple trials (see captions).
We do the same experiments for both the payload and no payload scenarios, with the only difference being the speed. For the scenario with no payload the maximum velocity was 2 m/s, while with the payload the maximum velocities were around 1.75 m/s.

\subsection{Results}
We compare our methods against the standard INDI controller and a geometric controller \citep{lee2010} with no residual force predictions.

\subsubsection{Importance of RPM Measurements for INDI}
In order to prove the reliance on RPM measurements of INDI, we test the performance of an adjusted version of INDI which uses pulse width modulation (PWM) values instead of RPM to measure the total force output of the rotors. For this experiment we train LINDI and NA-INDI on the predictions made by INDI-PWM, with results shown in Table \ref{tab:resultspwm}. PWM measurements are noisy and introduce delay-induced mismatch in thrust estimation.

The results show that using LINDI and NA-INDI to learn the residual force calculations from the PWM-based INDI greatly improves flight performance compared to the PWM-based INDI but still underperformed when compared to the basic Lee controller with no residual prediction. This highlights INDI's dependence on accurate sensor measurements. In addition, we observe that the RPM measurements are still required for training our learning methods. The improvement over INDI-PWM with LINDI-PWM and NA-INDI-PWM indicate that the denoising capabilities of LINDI can potentially improve the performance of the standard INDI.

\begin{table*}[!t]
  \centering
  \caption{Comparison of average deviation from desired flight path (centimeters) of the quadrotor with no payload using INDI based on thrust from the PWM signal instead of RPM. The performance of each method was averaged over ten trials.}
  \begin{tabular}{|c|c|c|c|}
    \hline
    & Circle & Figure8 \\
    \hline
    Lee & $7.52 \pm 1.22$ & $7.29 \pm 0.31$ \\
    \hline
    INDI-PWM & $18.27 \pm 0.97$ & $18.89 \pm 1.82$ \\
    \hline
    LINDI-PWM & $15.33 \pm 1.90$ & $\mathbf{14.53} \pm 1.75$ \\
    \hline
    NA-INDI-PWM & $\mathbf{14.31} \pm 1.91$ & $16.71 \pm 2.80$ \\
    \hline
  \end{tabular}
  \label{tab:resultspwm}
\end{table*}

\subsubsection{Multirotor with no payload}

\begin{figure}[t]
    \centering
    \begin{minipage}[t]{0.3\textwidth}
        \includegraphics[width=\textwidth]{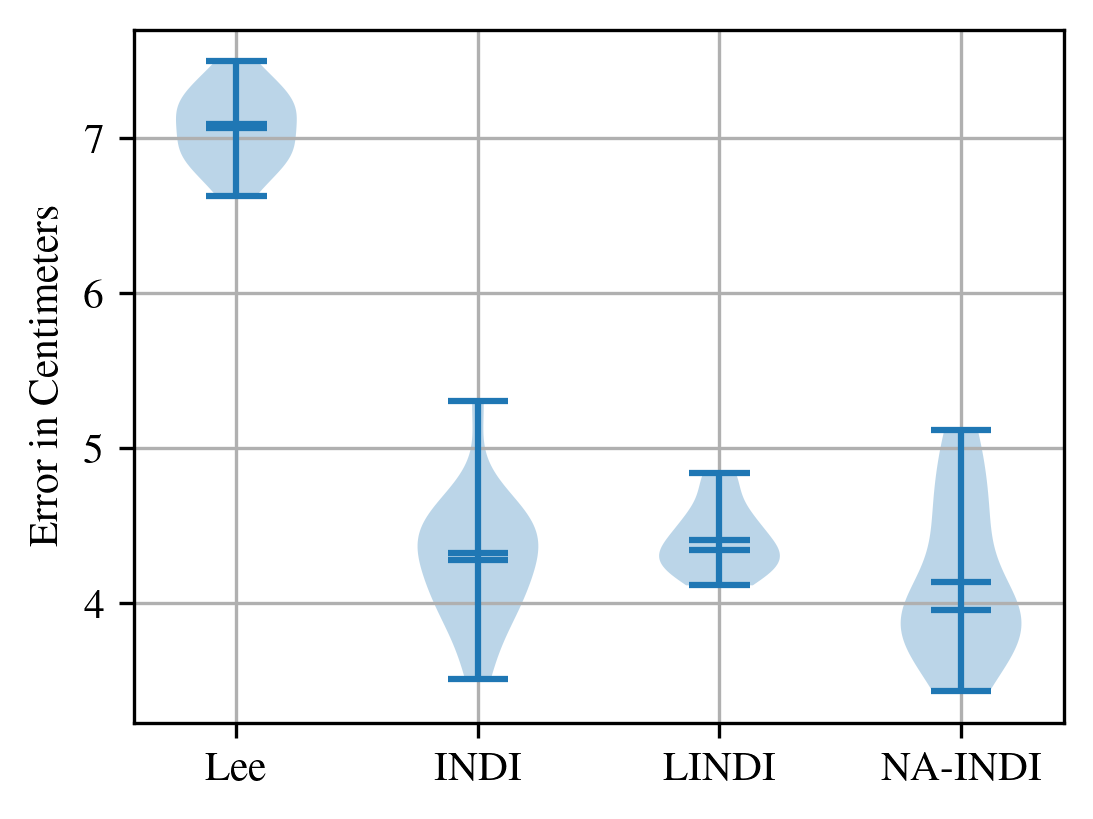}
        \caption*{(a) Figure8}
    \end{minipage}
    \hspace{1cm}
    \begin{minipage}[t]{0.3\textwidth}
        \includegraphics[width=\textwidth]{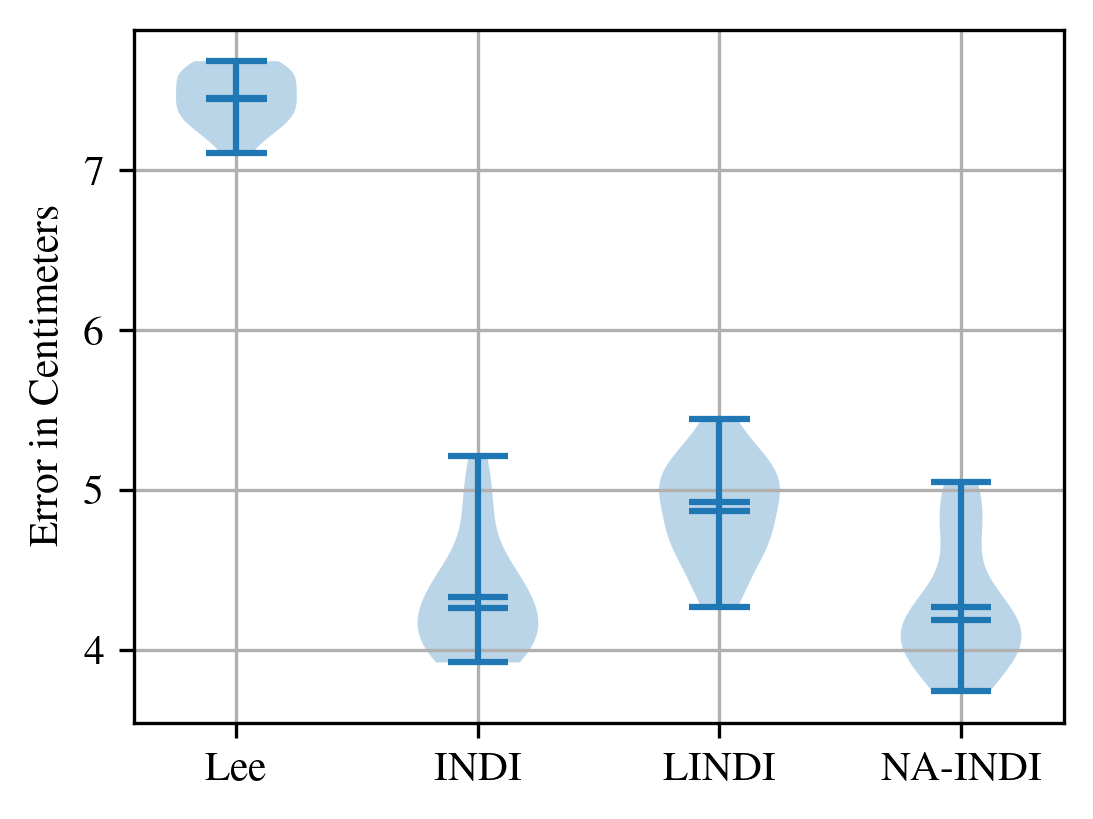}
        \caption*{(b) Circle}
    \end{minipage}
    \caption{Performance comparison between all four methods on the quadrotor with no payload for (a) the Figure8 flight trajectory and (b) the circle trajectory. Each violin plot has twenty datapoints: the mean tracking error (centimeters) of the multirotor position \eqref{eq:trackinError} for each trial. Mean values for Figure8: Lee: 7.07, INDI: 4.28, LINDI: 4.41, NA-INDI: \textbf{4.14}. Mean values for Circle: Lee: 7.45, INDI: 4.33, LINDI: 4.87, NA-INDI: \textbf{4.26}.}
    \label{fig:resultsvionopay}
\end{figure}

Figure \ref{fig:resultsvionopay} shows that residual prediction can reduce the tracking error to the desired flight trajectory by around 40\%. As can be seen from the results, LINDI leads to similar performance to INDI without the use of the RPM measurements, however, for the circular trajectory, we observe a slight drop in performance, likely because the trajectory is out of distribution relative to the training data. The results also show that the combination of sensor measurements plus learned residuals (NA-INDI) leads to the best performance, although with a small margin. This results from the noise in the RPM measurements being easier to filter, as the INDI component only needs to account for a smaller portion of the residuals, with the remaining effects handled by the LINDI component.


\subsubsection{Multirotor with Payload}

\begin{figure}[t]
    \centering

    \begin{minipage}[t]{0.3\textwidth}
        \centering
        \includegraphics[width=\textwidth]{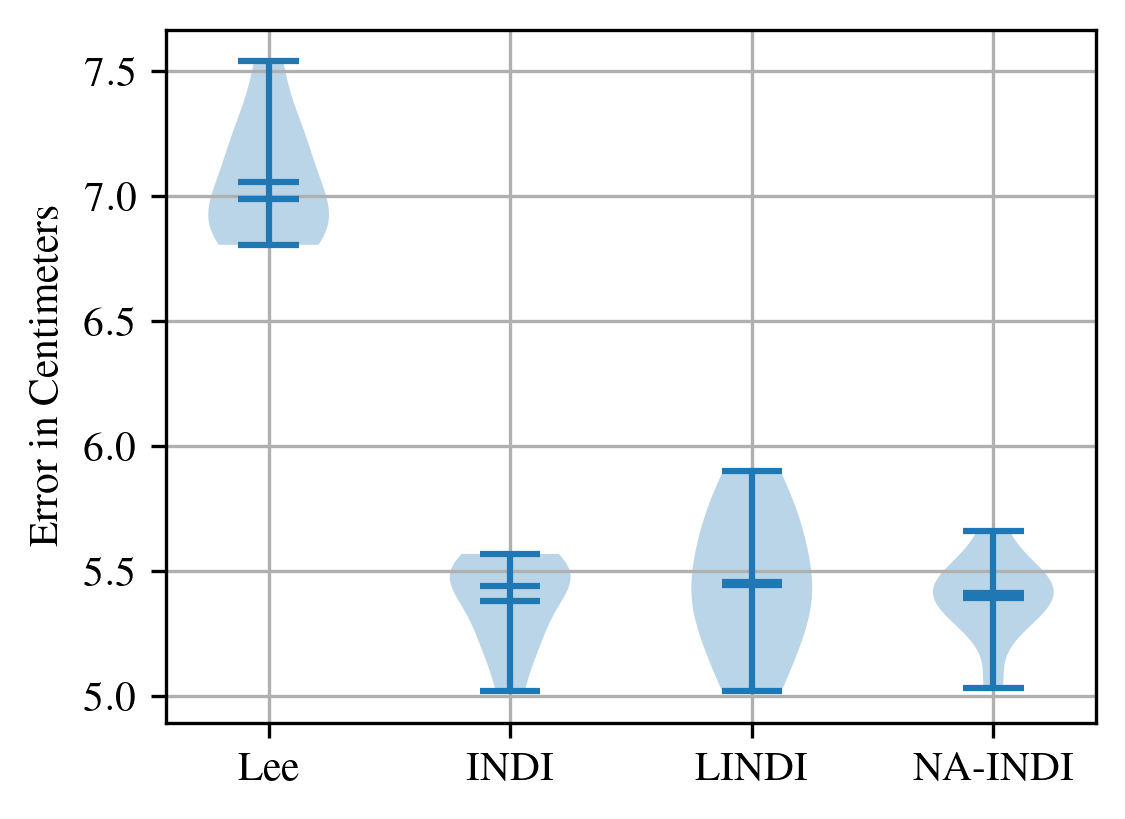}
        \caption*{(a) Figure8}
    \end{minipage}
    \hspace{1cm}
    \begin{minipage}[t]{0.3\textwidth}
        \centering
        \includegraphics[width=\textwidth]{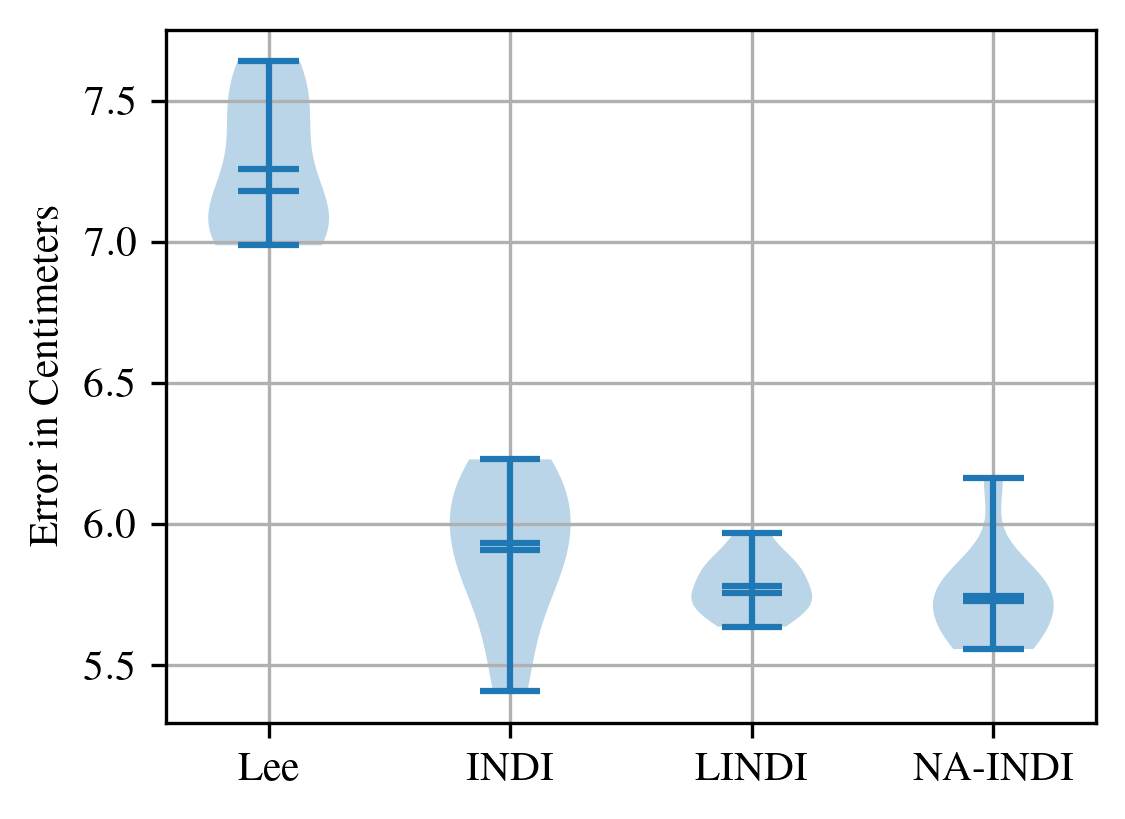}
        \caption*{(b) Circle}
    \end{minipage}

    \caption{Performance comparison between all four methods on the quadrotor carrying a payload for (a) the Figure8 flight trajectory and (b) the circle trajectory. Each violin plot has ten datapoints: the mean tracking error (centimeters) of the payload position \eqref{eq:trackinError} for each trial. Mean values for Figure8: Lee: 7.06, INDI: \textbf{5.38}, LINDI: 5.46, NA-INDI: 5.39. Mean values for Circle: Lee: 7.26, INDI: 5.91, LINDI: 5.78, NA-INDI: \textbf{5.74}.}
    \label{fig:resultsviopay}
\end{figure}

In Figure \ref{fig:resultsviopay} we can see similar results to the single quadrotor experiments. Once again LINDI is able to match the performance of INDI without the need of RPM measurements, even with the more complex dynamics of the payload. NA-INDI shows the potential to outperform INDI, but still remains similar in performance.


\section{Conclusion}
We introduce LINDI, a novel approach for learning residual predictions which trains a neural network on smooth data based on spline fitting. The results demonstrate that INDI can be effectively replaced by a neural network, which can output smoother residual force predictions without requiring special sensor measurements. While the proposed method requires a quadrotor equipped with RPM measurement sensors to collect the training data, once the neural network has been trained, it can be deployed on any number of quadrotors with similar specifications. This can significantly reduce the cost of maintaining a large fleet of drones with accurate flight controllers.

This work establishes a foundation for extending learning-based residual modeling toward more realistic scenarios involving external aerodynamic disturbances. 
Exploring how LINDI performs in the presence of such external effects offers a promising avenue for future research and could enable robust, disturbance-aware residual prediction for real-world operations.

\acks{The work was supported by the Deutsche Forschungsgemeinschaft (DFG, German Research Foundation) - 448549715.}
\bibliography{references}

\newpage
\appendix
\setcounter{section}{0}

\section{Cable-Suspended Payload Dynamics}
The dynamics of a single quadrotor transporting a point-mass payload via a cable modeled as a rigid rod are governed by the following equations.  
According to \cite{tang2014aggressive, sreenath2013geometric1uav, yu2020geometric, li2023rotortm}, the coupled quadrotor–payload dynamics can be expressed from Hamilton's principle as
\begin{align}
    \label{eq:dy}
    &m_T\mathbf{a}_p = \mathbf{u}^{\parallel} - ml\|\boldsymbol{\Omega}\|^2\vq - m_T g\mathbf{e}_z + \mathbf{f}_{a_p}, \nonumber \\
    &\dot{\vq} =  \boldsymbol{\Omega} \times \mathbf{q}, \nonumber \\
    &\dot{\boldsymbol{\Omega}} = -\frac{1}{ml}\vqhat_1\mathbf{u}^{\perp} + \mathbf{f}_{a_c}, \nonumber \\
    &\dot{\mR} = \mR\hat{\boldsymbol\omega}, \nonumber \\
    &\mJ\dot{\boldsymbol\omega} = \mJ\boldsymbol\omega\times\boldsymbol\omega + \boldsymbol{\tau}_u + \boldsymbol{\tau}_a,
\end{align}
where $m_T = m_p + m$ is the total mass of the payload–quadrotor system. 
The vectors $\mathbf{p}_p, \mathbf{v}_p, \mathbf{a}_p \in \mathbb{R}^3$ denote the position, velocity, and acceleration of the payload, respectively. 
The unit vector $\vq$ points from the quadrotor position $\mathbf{x}\in\mathbb{R}^3$ toward the payload position $\mathbf{x}_p$, and $\boldsymbol{\Omega}\in\mathbb{R}^3$ represents the angular velocity of the cable.  

The total control force applied by the quadrotor is decomposed as
\begin{equation}
\mathbf{u}_0 = \mathbf{u}^{\parallel} + \mathbf{u}^{\perp},
\end{equation}
where the parallel and perpendicular components are defined as
\begin{equation}
\mathbf{u}^{\parallel} = \vq\vq^{\top} f_u\mR\mathbf{e}_z, \qquad 
\mathbf{u}^{\perp} = -\hat{\vq}^2 f_u\mR\mathbf{e}_z,
\end{equation}
and $f_u$ is the collective thrust generated by the quadrotor.  

The residual aerodynamic forces acting on the system are represented by $\mathbf{f}_{a_p}$ and $\mathbf{f}_{a_c}$, applied to the payload and cable dynamics, respectively. 
The last two equations in \eqref{eq:dy} describe the rotational dynamics of the quadrotor.

\section{Cable-Suspended Payload Geometric-INDI Controller}
This section presents the complete mathematical derivation of an alternative geometric controller in ~\cite{yu2020geometric} and INDI formulation (our contribution) that explicitly includes the cable dynamics.
The purpose is to provide the theoretical foundation complementing the INDI variant used in the main paper.

We adopt a geometric INDI formulation in which the payload tracking is regulated by a virtual force controller $\mathbf{F}_d$ along the cable direction, while the cable attitude is controlled by a perpendicular component. 
We consider a smooth reference trajectory for the payload, defined by its desired position, velocity, acceleration, and higher-order derivatives up to the $n^{\text{th}}$ order $(\mathbf{p}_{p_d}, \mathbf{v}_{p_d}, \mathbf{a}_{p_d}, \mathbf{j}_{p_d}, \mathbf{x}^{(n)}_{p_d})$. Let
\begin{equation}
    \mathbf{e}_p = \mathbf{p}_{p_d} - \mathbf{p}_p, 
    \qquad 
    \mathbf{e}_v =  \mathbf{v}_{p_d} - \mathbf{v}_p,
\end{equation}
and let the feedback gains be $k_p,k_v>0$ (scalars) or, more generally, $\mathbf{K}_p,\mathbf{K}_v \in \mathbb{R}^{3\times 3}$ positive definite (diagonal) matrices. The desired payload virtual force is
\begin{equation}
\label{eq:F_d}
    \mathbf{F}_d 
    = m_T(\mathbf{a}_{p_d} + \mathbf{K}_p \mathbf{e}_p + \mathbf{K}_v \mathbf{e}_v + g\,\ve_z).
\end{equation}
The parallel control component is then chosen as
\begin{equation}
\label{eq:u_parallel}
    \mathbf{u}^{\parallel} 
    = \vq\vq^\top \mathbf{F}_d + m l\|\boldsymbol{\Omega}\|^2 \vq - \mathbf{f}_{a_p},
\end{equation}

To regulate the cable direction, define the desired direction
\begin{equation}
\label{eq:q_d}
    \vq_d = -\frac{\mathbf{F}_d}{\|\mathbf{F}_d\|},
\end{equation}
and let $\boldsymbol{\Omega}_d$ satisfy the kinematics $\boldsymbol{\Omega}_d = \dot{\vq}_d \times \vq_d$, where $\dot{\vq}_d$ is computed with numerical differentiation from $\vq_d$ or ideally from the differential flatness computation in \cite{tang2014aggressive} as
\begin{equation}
   \dot{\vq}_d =  \frac{-m_p\mathbf{j}_{p_d} + \dot{T}\vq_d}{\|m_p\mathbf{a}_{p_d}+g\mathbf{e}_z\|}, \quad  \dot{T} = -m_p(\mathbf{j}_{p_d}\cdot\vq_d).
\end{equation} 
Define the cable attitude and angular-rate errors (consistent with \cite{sreenath2013geometric1uav, yu2020geometric}) as
\begin{equation}
\label{eq:cable_errors}
    \mathbf{e}_q = \vq_d \times \vq,
    \quad
    \mathbf{e}_{\boldsymbol{\Omega}} = \boldsymbol{\Omega} + \vqhat^{2}\,\boldsymbol{\Omega}_d,
\end{equation}
and choose positive gains $k_{\vq},k_{\boldsymbol{\Omega}}>0$ (or $\mathbf{K}_{\vq},\mathbf{K}_{\boldsymbol{\Omega}}$ positive definite diagonal matrices). 
Using the cable dynamics in \eqref{eq:dy}, a perpendicular control that stabilizes $(\vq,\boldsymbol{\Omega})\!\to\!(\vq_d,\boldsymbol{\Omega}_d)$ is
\begin{equation}
\label{eq:u_perp}
\mathbf{u}^{\perp} 
= m l\,\vqhat_{1}(
   - \mathbf{K}_{\vq}\,\mathbf{e}_{q}
   - \mathbf{K}_{\boldsymbol{\Omega}}\,\mathbf{e}_{\boldsymbol{\Omega}}
   - (\vq \cdot \boldsymbol{\Omega}_{d})\,\dot{\vq}
   - \vqhat^{2}\dot{\boldsymbol{\Omega}}_{d}
   - \mathbf{f}_{a_c}),
\end{equation}

Finally, The collective thrust of the quadrotor is defined as $f = \mathbf{u}_0 \cdot \mathbf{R}\mathbf{e}_z$, and the attitude is tracked using the geometric controller proposed in \citet{lee2010}.

\subsection{INDI Formulation}
In the geometric INDI framework, the residual aerodynamic forces $\mathbf{f}_{a_p}$ and $\mathbf{f}_{a_c}$ correspond to the unmodeled disturbances acting on the payload and the cable, respectively, as defined in the dynamics equations in \eqref{eq:dy}. 
The term $\mathbf{f}_{a_p}$ captures the discrepancy between the measured acceleration of the payload $\ddot{\mathbf{x}}_p$ (left hand side) and the nominal acceleration predicted by the model (right hand side). Similarly, $\mathbf{f}_{a_c}$ is derived analogously from the measured angular acceleration of the cable $\dot{\boldsymbol{\Omega}}$. 
These residuals are used by the INDI controller to incrementally adjust the control inputs based on the difference between the measured and model-predicted dynamics as shown in Eqs. \eqref{eq:u_parallel} and \eqref{eq:u_perp}.

\subsubsection{Practical Implementation}
While the previous section provides the theoretical derivation, the practical realization of this formulation presents significant challenges.
We outline below the sensing and numerical differentiation requirements, along with the sources of estimation noise that make the real-time implementation difficult on lightweight multirotors hardware.
In our experimental setup, the payload position $\mathbf{p}_p$ is measured directly from the motion capture (MoCap) system. 
The payload acceleration $\mathbf{a}_p$ is estimated by twice numerically differentiating the measured position data, with a Butterworth filter applied after each differentiation step to obtain smooth velocity and acceleration estimates.

Similarly, the cable direction $\vq$ is computed from the measured positions of the quadrotor and the payload. 
The cable angular velocity $\boldsymbol{\Omega}$ and angular acceleration $\dot{\boldsymbol{\Omega}}$ are estimated through numerical differentiation of $\vq$ and filtered using the same Butterworth approach (applied after each step). 
In practice, this introduces a trade-off between signal noise and measurement delay: a higher filter cutoff frequency yields a noisier but more responsive signal, while a lower cutoff produces smoother estimates at the cost of additional delay.

\textit{Note:} In principle, one might implement an Extended Kalman Filter (EKF) for the payload and the cable state as in \cite{2025-sun-AgileCooperativeAerial}. However as the payload dynamics are tightly coupled with those of the quadrotor and then the estimation leads to large matrix inversions, which are computationally demanding on the STM32 firmware of the Crazyflie platform.

\subsubsection{Discussion}

Building on the practical implementation discussed above, we now reflect on the limitations and potential directions for improving the INDI formulation.
In principle, the proposed INDI formulation was not experimentally validated in this work, as the estimated states were affected either by significant noise introduced through numerical differentiation or by the delays introduced when using low Butterworth filter cut-off frequencies.
In particular, the estimation of the cable angular acceleration $\dot{\boldsymbol{\Omega}}$ proved highly sensitive to measurement noise, even after filtering, which led to residual force estimates that were too noisy to yield reliable experimental results. 
Nevertheless, we include this formulation as it represents the theoretically consistent approach derived directly from the system dynamics.

In contrast, the INDI formulation used in the main paper represents a trade-off between modeling fidelity and practical feasibility. 
By dropping the explicit cable dynamics and relying on IMU-measured accelerations of the multirotors (which already capture the coupling effects through the cable tension) the controller assumes perfect knowledge of these effects while gaining smoother residual estimates and improved real-time stability.

A promising direction to address these challenges is to learn the residual dynamics directly using a data-driven model, where temporal input histories implicitly capture higher-order state information, thereby reducing the reliance on explicit state estimation.
This analysis provides the foundation for future work on integrating learned residual models and state estimation methods for reliable INDI-based control of cable-suspended systems.

\end{document}